\documentclass{article}

\usepackage{microtype}
\usepackage{graphicx}
\usepackage{subfigure}
\usepackage{booktabs} 

\usepackage{hyperref}

\usepackage[accepted]{icml2025}

\usepackage{amsmath}
\usepackage{amssymb}
\usepackage{mathtools}
\usepackage{amsthm}
\usepackage{array, booktabs}
\usepackage{arydshln}

\usepackage[capitalize,noabbrev]{cleveref}

\theoremstyle{plain}

\theoremstyle{definition}

\theoremstyle{remark}

\usepackage[textsize=tiny]{todonotes}

\icmltitlerunning{Towards a Foundation Model for  Communication Systems}

\begin{document}

\twocolumn[
\icmltitle{Towards a Foundation Model for  Communication Systems}



\icmlsetsymbol{equal}{*}

\begin{icmlauthorlist}
\icmlauthor{Davide Buffelli}{equal,yyy}
\icmlauthor{Sowmen Das}{equal,yyy}
\icmlauthor{Yu-Wei Lin}{equal,xxx}
\icmlauthor{Sattar Vakili}{equal,yyy}
\icmlauthor{Chien-Yi Wang5}{xxx}
\icmlauthor{Masoud Attarifar}{xxx2}
\icmlauthor{Pritthijit Nath}{zzz}
\icmlauthor{Da-shan Shiu}{yyy2}
\end{icmlauthorlist}

\icmlaffiliation{yyy}{MediaTek Research, London, UK}
\icmlaffiliation{yyy2}{MediaTek Research, Taipei, R.O.C.}
\icmlaffiliation{xxx}{MediaTek, Hsinchu, R.O.C.}
\icmlaffiliation{xxx2}{MediaTek, Cambourne, UK}
\icmlaffiliation{zzz}{University of Cambridge, Cambridge, UK}

\icmlcorrespondingauthor{Davide Buffelli}{davide.buffelli@mtkresearch.com}
\icmlcorrespondingauthor{Sattar Vakili}{sattar.vakili@mtkresearch.com}

\icmlkeywords{Machine Learning, ICML}

\vskip 0.3in
]



\printAffiliationsAndNotice{\icmlEqualContribution} 

\begin{abstract}
Artificial Intelligence (AI) has demonstrated unprecedented performance across various domains, and its application to communication systems is an active area of research. While current methods focus on task-specific solutions, the broader trend in AI is shifting toward large \emph{general} models capable of supporting multiple applications. In this work, we take a step toward a \emph{foundation model for communication data}---a transformer-based, multi-modal model designed to operate directly on communication data. We propose methodologies to address key challenges, including tokenization, positional embedding, multimodality, variable feature sizes, and normalization. Furthermore, we empirically demonstrate that such a model can successfully estimate multiple features, including transmission rank, selected precoder, Doppler spread, and delay profile.
\end{abstract}

\section{Introduction}
In modern communication systems, numerous estimation and optimization tasks are required to ensure effective signal transmission and reception. Some examples include channel estimation, Doppler spectrum and delay spread estimation, and channel state information (CSI) feedback, which includes precoding matrix and its rank.

In the past few years, an increasingly large amount of research has focused on replacing traditional solutions to the above tasks with machine learning, and deep learning in particular \cite{9550825,THAKKAR2020119}. These methods are based on the idea of learning from data rather than relying on handcrafted heuristics, paving the way for more intelligent and efficient networks. Indeed, AI is already expected to become a key component of 6G networks \cite{10.1109/MWC.008.2200157} and will play an even more central role in future communication systems \cite{saad2024artificialgeneralintelligenceaginative}.

Current research in this domain has focused on designing task-specific machine learning solutions. These approaches involve collecting large amounts of supervised data, which is costly and time-consuming, and require a dedicated model for each task. In contrast, the broader field of AI is increasingly adopting a paradigm in which a single large model, referred to as a \emph{foundation model}, is trained on vast amounts of unsupervised (and thus easier to obtain) data. Such a foundation model learns intricate relationships in the data, and is then leveraged to perform downstream tasks, by either applying it directly, specializing it with some additional fine-tuning, or by using its learned representations. 

In this work, we move the first steps towards a \emph{Foundation Model for Communication Systems}. This is a large deep learning model trained on vast quantities of unsupervised data coming from network communication.
While foundation models have been proposed for domains like natural language, code, images, physical data, and time series, communication systems present unique challenges that are not addressed by the current literature. In more detail, dealing with multiple features at each slot in the input sequence, features from different domains (categorical, integer, real and complex valued, matrices), and with varying sizes and magnitudes, requires carefully designed ad-hoc procedures and components.
We summarize our contributions as follows:
\begin{itemize}
    \item We design and train a foundation model for communication systems addressing key challenges including tokenization,
    positional embedding, multimodality, features of varying size, and
    normalization. 
    \item We design a simulation system to generate unsupervised data for the training of our foundation model.
    \item We test our foundation model in several scenarios, showing  it can successfully perform multiple estimations.
    \item {We report experimental results on how model size and dataset size affect estimation performance, providing initial insights into the scaling behavior of foundation models for communication systems (see Figures~\ref{fig:scaling_data_gen} and~\ref{fig:scaling_training}).}
\end{itemize}

\section{Related Work}
We focus our discussion on foundation models, and large language models (LLMs) applied to communication applications. The literature on machine learning for communication systems is rapidly growing, and a complete presentation would be out of the scope of this paper. We refer the interested reader to recent surveys on machine learning  \cite{s24061925,wang2024machinelearningcommunicationsroad,s23218792}, and deep learning in particular \cite{9914567,Liao_2020}, in this domain.

\subsection{Foundation Models}
The term \emph{foundation model} was introduced in the context of LLMs \cite{Bommasani2021FoundationModels}, i.e., models trained on large quantities of textual data.  Multimodal foundation models that combine data from different modalities have also been introduced. The most popular such models combine text with images \cite{alayrac2022flamingo,chen2025expandingperformanceboundariesopensource,dai2024nvlmopenfrontierclassmultimodal,geminiteam2024geminifamilyhighlycapable,openAIo1}.

Recently, significant effort has been dedicated to creating foundation models  outside of the traditional textual, visual, or audio domains. Examples of this are foundation models for signals from wearable devices \cite{narayanswamy2025scaling,abbaspourazad2024largescale}, earth-related signals \cite{bodnar2024foundationmodelearth}, and time-series \cite{goswami2024moment,rasul2024lagllama,shi2025timemoe,das2024decoderonlyfoundationmodeltimeseries,liu2024moiraimoeempoweringtimeseries,ekambaram2024tiny,liu2025sundialfamilyhighlycapable}.

While these models share commonalities to our work, communication data poses unique challenges that are not addressed in current literature. In particular, prior works focus on settings with few features and low heterogeneity—unlike communication systems, as discussed in Section~\ref{sec:method}.

\subsection{LLMs for Communication Systems}
With the rise in popularity of LLMs, several works have focused on introducing them into the context of communication systems. \cite{10582827} presents the challenges that need to be addressed to make existing pre-trained LLMs able to understand data from communication system. \cite{nikbakht2024tspecllmopensourcedatasetllm} introduce a dataset containing numerous technical documents that can be used to introduce knowledge related to communication systems to LLMs.
Finally, \cite{piovesan2024telecomlanguagemodelslarge} shows that ``small'' language models enhanced with Retrieval-Augmented Generation (RAG) \cite{10.5555/3495724.3496517} approaches can already understand communications data, and \cite{bornea2024telcoragnavigatingchallengesretrievalaugmented} further improve the RAG procedure for specialized technical documents.
These works aim to enrich LLMs with communication-related knowledge but do not operate on low-level data. In contrast, our model directly processes raw, heterogeneously structured communication data to perform effective feature estimation, without relying on external knowledge sources such as natural language.

\section{Communication System Model and Simulator}
\label{sec:comms}

In this section, we describe the communication system model and the simulated data used to train the foundation model.

\subsection{Communication System Model}

We consider a multiple-input multiple-output (MIMO) orthogonal frequency-division multiplexing (OFDM) system of $N_T$ transmit (Tx) antennas, $N_R$ receive (Rx) antennas, $K$ subcarriers per OFDM symbol with subcarrier spacing $f_\text{sc}$ Hz, and $L$ OFDM symbols per slot. We restrict our attention to channel state information (CSI) acquisition, which is an important subsystem of wireless communications. For CSI acquisition, the transmitter sends pilot signals to the receiver. The receiver estimates the CSI and then reports to the transmitter.

The input--output in the frequency domain are related as 
\begin{equation*}
    \mathbf{y}[k,l,n]=\mathbf{H}[k,l,n]\mathbf{x}[k,l,n] + \mathbf{z}[k,l,n],
\end{equation*}
where $\mathbf{y}[k,l,n]\in\mathbb{C}^{N_R}$ is the Rx signal vector at subcarrier $k$ and symbol $l$ in slot $n$, $\mathbf{x}[k,l,n]\in\mathbb{C}^{N_T}$ is the Tx signal vector, $\mathbf{H}[k,l,n]\in\mathbb{C}^{N_R\times N_T}$ is the channel matrix drawn from a wide-sense stationary random process of mean zero and unit power per entry, and $\mathbf{z}[k,l,n]\in\mathbb{C}^{N_R}$ is the additive noise independent of $\mathbf{H}$. We assume $\{\mathbf{z}[k,l,n]\}$ are independent and identically distributed (i.i.d.) circularly-symmetric complex Gaussian $\mathcal{CN}(\mathbf{0},\mathbf{C}_\text{n})$, where $\mathbf{0}$ denotes an all-zero vector and $\mathbf{C}_\text{n}$ denotes the noise covariance matrix. For data transmission, each modulated symbol vector $\mathbf{s}\in\mathcal{A}^{R}$ is precoded by $\mathbf{W}_{(R)}\in\mathcal{W}_R\subset\mathbb{C}^{N_T\times R}$, where $\mathcal{A}$ is the set of constellation points, $R$ denotes the rank (number of layers), and $\mathcal{W}_R$ is the codebook of rank $R$ known by the transmitter and the receiver. We assume that $\mathbb{E}[\|\mathbf{s}\|^2]=P$ and $\|\mathbf{W}_{(R)}\|_F^2=1$ for all $R$, where $\|\cdot\|_F$ denotes the Frobenius norm. We assume that the same precoder is applied to all subcarriers and symbols within a slot. Then, we have $\mathbf{x}[k,l,n]=\mathbf{W}_{(R)}[n]\mathbf{s}[k,l,n]$ and thus the input--output relation can be expanded as 
\begin{equation*}
    \mathbf{y}[k,l,n]=\mathbf{H}[k,l,n]\mathbf{W}_{(R)}[n]\mathbf{s}[k,l,n] + \mathbf{z}[k,l,n].
\end{equation*}
  
We divide the $K$ subcarriers into $B$ subcarrier groups of size $M$, so that $K = BM$. Denote $\mathcal{P}=\{(Mm,l): m\in\{1,\cdots,B\}, l\in\mathcal{S}\}$, where $\mathcal{S}$ denotes the set of symbols with pilots. The performance of CSI acquisition can be assessed by spectral efficiency defined as  
\begin{equation*}
\resizebox{\columnwidth}{!}{$
G(R,\mathbf{W}_{(R)})
= \frac{1}{|\mathcal{P}|}\sum_{(k,l)\in\mathcal{P}}\log\det\left(\mathbf{C}_\text{n}+P\mathbf{H}[k,l]\mathbf{W}_{(R)}\mathbf{W}^H_{(R)}\mathbf{H}^H[k,l]\right),
$}
\end{equation*}
which follows the capacity of MIMO channel with CSI at receiver (see e.g., Section 8.2.1 of \cite{Tse_book}).

To measure the channel matrices $\{\mathbf{H}[k,l]\}$, we transmit pilots following a comb-type structure: For each Tx antenna $j$, we set $\mathbf{x}[Mm+j,l,n]=\sqrt{P}\mathbf{e}_j$, where $l\in\mathcal{S}$ and $\mathbf{e}_j$ denotes the standard unit vector with one on the $j$-th position and zero otherwise. Besides, the noise covariance $\mathbf{C}_\text{n}$ needs to be estimated, so we set $\mathbf{x}[Mm+j,l,n]=\mathbf{0}$ for $j=N_T+1,\cdots,N_T+N_Z$ and assume $N_Z = M - N_T$.

The receiver needs to determine $(R,\mathbf{W}_{(R)})$ from the received signals at pilot subcarriers, also known as channel frequency response (CFR): For $j\in\{1,\cdots,N_T\}$,
\begin{equation*}
\begin{aligned}
\tilde{\mathbf{h}}_j[m,l,n] &\triangleq \mathbf{y}[Mm+j,l,n] \\
&= \sqrt{P}\mathbf{h}_j[Mm+j,l,n] + \mathbf{z}[Mm+j,l,n],
\end{aligned}
\end{equation*}
where $\mathbf{h}_j$ is the column $j$ of $\mathbf{H}$, and for $k\in\{1,\cdots,BN_Z\}$, 
\begin{equation*}
\resizebox{\columnwidth}{!}{$
\tilde{\mathbf{z}}[k,l,n] \triangleq \mathbf{z}[M\lfloor (k-1)/N_Z\rfloor+((k-1) \bmod N_Z)+1,l,n].
$}
\end{equation*}
To simplify notation, we assume per-slot operation and hence omit the slot index $n$ in the remainder of this section.

\subsubsection{Noise covariance estimation}
The noise covariance can be estimated by 
\begin{equation*}
\hat{\mathbf{C}}_\text{n}=\frac{1}{BN_Z|\mathcal{S}|}\sum_{k=1}^{BN_Z}\sum_{l\in\mathcal{S}}\tilde{\mathbf{z}}[k,l]\tilde{\mathbf{z}}^H[k,l].
\end{equation*}
The average noise power at Rx antenna $i$ is denoted as $\hat{\sigma}_i^2\triangleq[\hat{\mathbf{C}}_\text{n}]_{ii}$.

\subsubsection{Signal power estimation}
The signal power $P$ can be estimated as 
\begin{equation*}
\hat{P} =\frac{1}{N_TB|\mathcal{S}|N_R}\sum_{j=1}^{N_T}\sum_{m=1}^{B}\sum_{l\in\mathcal{S}}\left\|\tilde{\mathbf{h}}_j[m,l]\right\|^2 - \frac{1}{N_R}\sum_{i=1}^{N_R}\hat{\sigma}_i^2.
\end{equation*}

It is desirable to perform channel denoising before determining the rank and precoder. We perform robust channel estimation by assuming the power delay profile and the Doppler spectrum are both of rectangular shape \cite{Li_robustCE}. 

\subsubsection{Delay profile estimation}
For robust channel estimation in frequency domain, we assume that any two channel gains separated by $\Delta$ pilots in the same OFDM symbol, say $h_{(f),m}$ and $h_{(f),m+\Delta}$, satisfy that 
\begin{equation*}
\begin{aligned}
&\frac{\mathbb{E}[h_{(f),m+\Delta}h_{(f),m}^*]}{\sqrt{\mathbb{E}[|h_{(f),m+\Delta}|^2]\mathbb{E}[|h_{(f),m}|^2]}} \\
&= e^{-j2\pi\mu\Delta Mf_\text{sc}}\frac{\sin(\pi \ell\Delta Mf_\text{sc})}{\pi \ell\Delta Mf_\text{sc}},
\end{aligned}
\end{equation*}
where $\mu$ and $\ell$ are the center and the length of the delay profile, respectively. The center and the length are estimated by the following procedure: For the tuple of Tx antenna $j$, Rx antenna $i$, and symbol $l$, we collect its associated CFR into a vector $\tilde{\mathbf{h}}_{(f)}[i,j,l]$ and transform the CFR into the channel impulse response (CIR) $\tilde{\mathbf{h}}_{(t)}[i,j,l]$ through inverse discrete Fourier transform (IDFT) after zero padding to size $N_\text{FFT}$, which is the smallest power of 2 larger than $K$. The noisy delay profile $\tilde{\mathbf{p}}_{(t)}[i]$ at Rx antenna $i$ is then calculated as 
\begin{equation*}
\tilde{\mathbf{p}}_{(t)}[i] = \frac{1}{N_T|\mathcal{S}|}\sum_{j=1}^{N_T}\sum_{l\in\mathcal{S}} \left|\tilde{\mathbf{h}}_{(t)}[i,j,l]\right|^2.
\end{equation*}
Only the tap(s) with power larger than $3\hat{\sigma}_i^2$ are considered as including the desired signal. Denote $\mathcal{D}[i]=\left\{n\big|\left[\tilde{\mathbf{p}}_{(t)}[i]\right]_n>3\hat{\sigma}_i^2, 0\le n < N_\text{FFT}\right\}$. Then, the starting and ending positions of the delay profile $(\hat{n}_\text{start}[i],\hat{n}_\text{end}[i])$ are found as the pair that covers $\mathcal{D}[i]$ with the minimum length, considering invariance under circular shift.
Finally, we have 
\begin{equation*}
\begin{aligned}
\hat{\mu}[i] &= \frac{1}{Mf_\text{sc}N_\text{FFT}}\left(\frac{\hat{n}_\text{start}[i]+\hat{n}_\text{end}[i]}{2}\right),\\ 
\hat{\ell}[i] &= \frac{1}{Mf_\text{sc}N_\text{FFT}}(\hat{n}_\text{start}[i]-\hat{n}_\text{end}[i]+1),
\end{aligned}
\end{equation*}
and the estimated correlation matrix $\hat{\mathbf{R}}_\text{f,robust}$ with
\begin{equation*}
\begin{aligned}
&[\hat{\mathbf{R}}_\text{f,robust}]_{m_1,m_2} \\
&=e^{-j2\pi\hat{\mu}[i](m_1-m_2) Mf_\text{sc}}\frac{\sin(\pi \hat{\ell}[i](m_1-m_2) Mf_\text{sc})}{\pi \hat{\ell}[i](m_1-m_2) Mf_\text{sc}}.
\end{aligned}
\end{equation*}
In case that $\mathcal{D}[i]$ is empty, we set $\hat{\mu}[i]=\frac{\arg\max_n [\tilde{\mathbf{p}}_{(t)}[i]]_n}{Mf_\text{sc}N_\text{FFT}}$ and $\hat{\ell}[i]=\frac{1}{Mf_\text{sc}N_\text{FFT}}$.
Following the above procedure, for each Rx antenna $i$, the genie center $\mu[i]$ and the genie length $\ell[i]$ for the noiseless delay profile $\mathbf{p}_{(t)}[i]$ can be calculated. With an abuse of notation, we denote by $\mathbf{R}_\text{f,robust}[i]$ the genie robust frequency correlation matrix.

\subsubsection{Doppler spectrum estimation}
For robust channel estimation in time domain, we assume that any two channel gains $h_{(t),l_1}$ and $h_{(t),l_2}$ of the same subcarrier in symbol $l_1$ and $l_2$, respectively, satisfy that 
\begin{equation*}
\frac{\mathbb{E}[h_{(t),l_1}h_{(t),l_2}^*]}{\sqrt{\mathbb{E}[|h_{(t),l_1}|^2]\mathbb{E}[|h_{(t),l_2}|^2]}} = \frac{\sin(\pi w(l_1-l_2)T}{\pi w(l_1-l_2)T},
\end{equation*}
where $w$ is the width of the Doppler spectrum and $T$ is the OFDM symbol duration. For the tuple of Tx antenna $j$, Rx antenna $i$, and subcarrier group $m$, we collect its associated received signals into a vector $\tilde{\mathbf{h}}_{(t)}[i,j,m]$. Then, the time covariance matrix of Rx antenna $i$ can be estimated as
\begin{equation*}
\hat{\mathbf{C}}_\text{time}[i] = \frac{1}{N_TB}\sum_{j=1}^{N_T}\sum_{m=1}^B \tilde{\mathbf{h}}_{(t)}[i,j,m]\tilde{\mathbf{h}}^H_{(t)}[i,j,m]-\hat{\sigma}_i^2\mathbf{I}_{|\mathcal{S}|}.
\end{equation*}
The corresponding estimated time correlation matrix is denoted by $\hat{\mathbf{R}}_\text{time}[i]$, which can be calculated from the entries of $\hat{\mathbf{C}}_\text{time}$ by $r_{ij}=c_{ij}/\sqrt{c_{ii}c_{jj}}$. Then, the width $w$ can be found by 
\begin{equation*}
\hat{w}[i] = \operatorname*{\arg\min}_{w}\|\mathbf{R}_\text{t,robust}(w)-\hat{\mathbf{R}}_\text{time}[i]\|_F^2,
\end{equation*}
where
\begin{equation*}
[\mathbf{R}_\text{t,robust}(w)]_{l_1,l_2} = \frac{\sin(\pi w(l_1-l_2)T}{\pi w(l_1-l_2)T}.
\end{equation*}
Denote $\hat{\mathbf{R}}_\text{t,robust}[i]=\mathbf{R}_\text{t,robust}(\hat{w}[i])$ the estimated robust time correlation matrix of Rx antenna $i$. The genie time covariance matrix $\mathbf{C}_\text{time}[i]$ is calculated by replacing $\tilde{\mathbf{h}}_{(t)}[i,j,m]$ and $\hat{\sigma}_i^2$ by $\mathbf{h}_{(t)}[i,j,m]$ and $0$, respectively. Following the above procedure, the genie width $w[i]$ and the genie time correlation matrix $\mathbf{R}_\text{time}[i]$ can be calculated. We denote by $\mathbf{R}_\text{t,robust}[i]=\mathbf{R}_\text{t,robust}(w[i])$ the genie robust time correlation matrix.

Once we have $\hat{\mathbf{R}}_\text{f,robust}[i]$ and $\hat{\mathbf{R}}_\text{t,robust}[i]$, the robust channel estimation is performed as follows:
For Rx antenna $i$ and Tx antenna $j$, 
\begin{equation*}
\begin{aligned}
&\left[\hat{\mathbf{h}}_f[i,j,l]\right]_{l\in\mathcal{S}} \\
&= \hat{\mathbf{R}}_\text{robust}[i]\left(\frac{\hat{\sigma}_i^2}{\hat{P}}\mathbf{I}+\hat{\mathbf{R}}_\text{robust}[i]\right)^{-1}\left[\tilde{\mathbf{h}}_f[i,j,l]\right]_{l\in\mathcal{S}}, 
\end{aligned}
\end{equation*}
where $\mathbf{R}_\text{robust}[i]=\mathbf{R}_\text{t,robust}[i]\otimes\mathbf{R}_\text{f,robust}[i]$, $\otimes$ is the Kronecker product, and $[\tilde{\mathbf{h}}_f(i,j,l)]_{l\in\mathcal{S}}$ denotes the concatenation of CFRs $\tilde{\mathbf{h}}_f[i,j,l]$ for all $l$ in $\mathcal{S}$.

\subsubsection{Precoder selection}
We denote by $\hat{\mathbf{H}}[m,l]$ the estimated channel matrix of subcarrier group $m$ and symbol $l$. The average whitened spatial covariance matrix is given by 
\begin{equation*}
\hat{\mathcal{C}}_{S} = \frac{1}{B|\mathcal{S}|}\sum_{m=1}^{B}\sum_{l\in\mathcal{S}}\hat{\mathbf{H}}^H[m,l]\hat{\mathbf{C}}_\text{n}^{-1}\hat{\mathbf{H}}[m,l].
\end{equation*}
For each rank $R$, the precoder $\mathbf{W}_{(R)}$ is selected as 
\begin{equation*}
\hat{\mathbf{W}}_{(R)} = \operatorname*{\arg\max}_{\mathbf{W}\in\mathcal{W}_R}\log\det\left(\mathbf{I}_{R}+\mathbf{W}^H\hat{\mathbf{C}}_S\mathbf{W}\right).
\end{equation*}

\subsubsection{Rank selection}
The reported rank $\hat{R}$ is selected as 
\begin{equation*}
\hat{R} = \operatorname*{\arg\max}_{R\in\{1,\cdots,N_R\}}\log\det\left(\mathbf{I}_{R}+\hat{\mathbf{W}}_{(R)}^H\hat{\mathbf{C}}_S\hat{\mathbf{W}}_{(R)}\right).
\end{equation*}

\subsection{Dataset}
\label{sec:dataset}
We use the open-source \textsc{Sionna} package~\cite{sionna}, together with the system model described in this section, to generate the dataset. At each time step $n$, corresponding to a slot, we record a selected set of diverse features to conduct proof-of-concept experiments. The features are listed in Table~\ref{table:feature-summary}.

\begin{table}[ht]
    \centering
    \caption{Selected features}
    \label{table:feature-summary}
    \vspace{2mm}
    \resizebox{\columnwidth}{!}{%
    \begin{tabular}{l|l}
        \toprule
        \textbf{Feature} & \textbf{Description} \\
        \midrule
        Channel type (categorical) & The type of \textsc{Sionna} channel (UMi, UMa, RMa)\\
        $K \in \mathbb{Z}^+$ & Number of subcarriers per OFDM symbol \\
        $\hat{\mathbf{C}}_\text{n} \in \mathbb{C}^{N_R \times N_R}$ & Estimated noise covariance matrix \\
        $\hat{\mathbf{R}}_\text{f, robust} \in \mathbb{C}^{B \times B}$ & Estimated frequency correlation matrix \\
        $\hat{\mathbf{C}}_\text{time} \in \mathbb{C}^{|\mathcal{S}| \times |\mathcal{S}|}$ & Estimated time covariance matrix \\
        $\hat{\mathbf{R}}_\text{time} \in \mathbb{C}^{|\mathcal{S}| \times |\mathcal{S}|}$ & Estimated time correlation matrix \\
        $\hat{\mu}, \hat{\ell} \in \mathbb{R}$ & Estimated center and length of delay profile \\
        $\hat{w} \in \mathbb{R}^+$ & Estimated width of Doppler spectrum\\
        $\hat{\mathbf{W}}_{(R)} \in \mathbb{C}^{N_T \times R}$ & Selected precoder of rank $R$ \\
        $\hat{R} \in \mathbb{Z}^+$ & Selected transmission rank \\
        $\hat{G} \in \mathbb{R}^+$ & Estimated Spectral Efficiency\\
        \bottomrule
    \end{tabular}
    }
\end{table}

The symbol duration and cyclic prefix length follow the fifth generation (5G) New Radio (NR) numerology~\cite{38211}. Specifically, we use a normal cyclic prefix. In NR, a slot consists of 14 OFDM symbols, hence $L = 14$. The configurations used to generate the dataset is listed in Table~\ref{table:config-summary}, wherein SNR is randomly drawn following uniform distribution. To reduce implementation complexity, the parameters related to $(\hat{w}, \hat{\mu}, \hat{l})$ are selected from predefined candidate sets. 

\begin{table}[ht]
    \centering
    \caption{Dataset configuration}
    \label{table:config-summary}
    \vspace{2mm}
    \resizebox{\columnwidth}{!}{%
    \begin{tabular}{l|l}
        \toprule
        \textbf{Parameter} & \textbf{Values} \\
        \midrule
        Channel type & UMi, UMa, RMa \\
        (Center frequency, subcarrier spacing) (Hz) & (2.6G, 15k),  (3.5G, 30k)\\
        SNR (dB) & [0, 30] \\
        Receiver speed (km/h) & 3, 10, 30, 60, 90 \\
        $(N_T, N_R)$ & (4, 1), (4, 2), (4, 4), (8, 1), (8, 2), (8, 4)\\
        Number of subcarrier group ($B$) & 25, 50, 75, 100 \\
        Size of subcarrier group ($M$)& 12, 24, 48 \\
        Set of pilot symbols ($\mathcal{S}$) & \{2, 8\}, \{2, 6, 10\}, \{4, 8, 12\}, \{2, 5, 8, 11\}\\
        \bottomrule
    \end{tabular}
    }
\end{table}

Considering all possible combinations for the values in Table \ref{table:config-summary}, there is a total of 86403 unique simulation settings. As generating data for all configurations would be intractable, we randomly select a subset of  29000 configurations. For each selected configuration we run the simulation 8 times, each time with a different SNR (sampled randomly with uniform distribution from the range specified in the above table) for 100 slots. Each simulation uses a unique random seed. We then divide the obtained data into 5-slot sequences, each one representing one datapoint to be used as input for our model.

\begin{figure*}[htbp]
\centerline{\includegraphics[width=0.78\textwidth]
{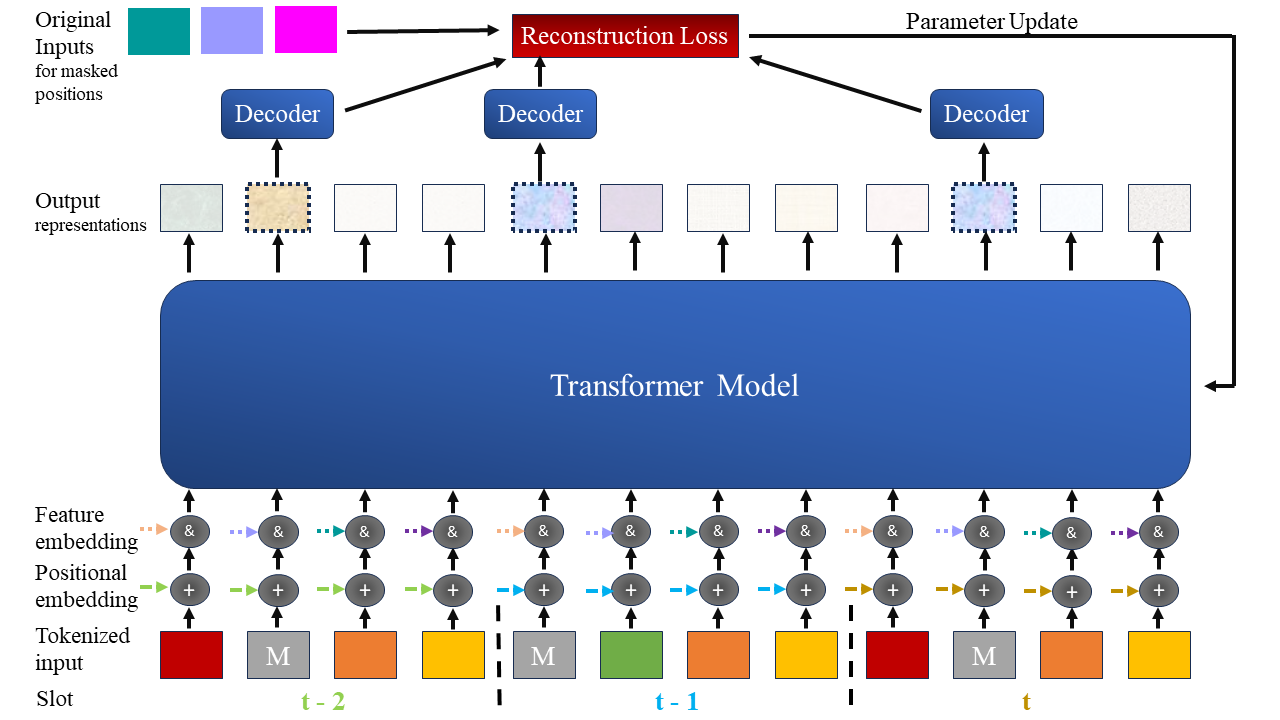}}
\caption{\textbf{Scheme of our foundation model for communications data and the pre-training procedure.} Tokens for each feature are first added to a positional embedding; features at the same input slot receive the same positional embedding (indicated by the color of the arrows). A feature embedding is then concatenated (\&); the tokens for the same feature at different slots receive the same feature embedding. The transformer processes these tokens and outputs a representation for each input. During pre-training, a subset of input features is masked (denoted as ``M''). The output representations corresponding to the masked inputs are decoded into the original feature space, and a reconstruction loss is computed. The model is trained to minimize this loss.}

\label{fig:model}
\end{figure*}

\section{Methodology}\label{sec:method}

The setting of communication systems presents several challenges that hinder the development of a foundation model. These systems are inherently complex and involve a large number of interdependent variables. Factors such as the choice of modulation, channel conditions (e.g., signal-to-noise ratio, multipath propagation, Doppler shift, fading), and the internal states of components like transmitters and receivers all influence system behavior. Moreover, the data in communication systems is highly heterogeneous. Some variables vary across consecutive transmissions over the same channel, while others remain constant for a given channel but change across different channels. Variables can be categorical, scalar, vector, or matrix-valued; their dimensionality may vary with system configuration, and they may belong to either the real or complex domain.

These characteristics require a foundation model for communication systems to reason over and capture relationships among a large number of diverse and heterogeneous variables. In the remainder of this section, we present our methodology for training such a model, covering data pre-processing, model design, and pre-training strategy.

\paragraph{Pre-processing}
Two important aspects must be considered.
The first is that for multi-dimensional features, the model must handle varying dimensionality depending on the current system settings. Since neural networks require fixed input sizes, we identify the maximum dimensionality for each feature and apply zero-padding to those with smaller dimensions. Padding masks are then introduced in the neural network computations to ensure that the model ignores the padded values.

The second aspect is that different features have different magnitudes, which can further vary significantly depending on the system configuration. We adopt a normalization strategy tailored to the properties of each feature. Scalar features are normalized using global statistics—specifically, we compute the mean and standard deviation over the training set and use them for normalization. Matrix and vector features are handled individually: we either normalize them by computing statistics across the input sequence or leave them unchanged (e.g., correlation matrices, which are already normalized). For the features that require padding, the normalization statistics are computed on the unpadded elements (and the normalization is applied only on those elements).

\paragraph{Tokenization} 
Given the heterogeneity of the features in our data, we divide them into four categories: scalars, vectors, matrices, and categorical. We design a tokenization method for each category as follows:

\begin{itemize}
    \item \textbf{Scalars:} we use Fourier encodings: 
    $\text{Tok}(x) = \left[ \cos \frac{2 \pi x}{\lambda_i}, \sin \frac{2 \pi x}{\lambda_i} \right]$ for $0 \le i < D/2$, 
    where the $\lambda_i$ are logarithmically spaced values taken from an interval $[\lambda_{\text{min}}, \lambda_{\text{max}}]$, chosen based on the scale and variability of the feature.

    \item \textbf{Vectors:} given a vector $x \in \mathbb{R}^d$, we apply a learnable linear transformation $W x$, where $W \in \mathbb{R}^{D \times d}$.

    \item \textbf{Matrices:} following the approach from ViT~\cite{dosovitskiy2021an}, we divide a matrix $X \in \mathbb{R}^{d_1 \times d_2}$ into ``patches'': $\{X^{(p)}_j \in \mathbb{R}^{p_1 \times p_2}, j=1, \dots, z \}$, where $p_1$ and $p_2$ are patch sizes, and $z = (d_1/p_1) \cdot (d_2/p_2)$ is the number of patches. Each patch is then flattened into a vector of size $p = p_1 \cdot p_2$ and transformed using a linear map $W \in \mathbb{R}^{D \times p}$. A matrix is thus converted into $z$ tokens, providing finer granularity to encode its structure. We define patch sizes using a simple heuristic: we choose the size such that the number of tokens per matrix is at most $64$, with a minimum patch size of $8 \times 8$. Alternatively, the patch size could be treated as a tunable hyperparameter, but we avoid this extra cost as our heuristic performs well in our experiments.

    \item \textbf{Categorical:} we assign a learnable vector $V \in \mathbb{R}^D$ to each category.
\end{itemize}

To handle complex-valued features, we first normalize and pad them as needed, and then convert them into two-channel vectors or matrices by stacking the real and imaginary parts.

\paragraph{Transformer Model}
For the base architecture of the transformer, we follow the Llama model~\cite{grattafiori2024llama3herdmodels}, as adopted by several state-of-the-art open-source models.

Typically, transformer-based foundation models receive a single quantity (and thus a single token) per each element of the input sequence. In our case, however, we have multiple features—and hence multiple tokens—at each slot in the input sequence. We assign the same positional embedding to the tokens for all features within the same slot. This setup, however, means that the model cannot, by default, distinguish tokens corresponding to the same feature across different slots, as it has no information about which token, within those for a given slot, is related to which feature. To address this, we introduce a \emph{feature embedding}: a learnable vector associated with each input feature, which is concatenated to the corresponding token to let the model recognize instances of the same feature across slots.

Once the tokens have been processed by the transformer, we obtain an embedding vector for each input token. To compute the loss for a given feature, we need to decode the embedding vector(s) back into the original space of that feature. We design the decoding mechanism to follow an ``inverse'' process to the one used for tokenization:

\begin{itemize}
    \item \textbf{Scalars}: a learnable vector $U \in \mathbb{R}^{1 \times D}$ maps the hidden representation to a scalar
    \item \textbf{Vectors}: a learnable matrix $Q \in \mathbb{R}^{d \times D}$ maps the hidden representation to the original dimensionality
    \item \textbf{Matrices}: the token for each patch is mapped back to its original dimensionality with a learnable linear transformation $E \in \mathbb{R}^{p_1 \cdot p_2 \times D}$. The patches are then reshaped into a matrix in $\mathbb{R}^{p_1 \times p_2}$ and concatenated back into the original shape for the feature
    \item \textbf{Categorical}: a learnable matrix $R \in \mathbb{R}^{n_c \times D}$ maps the hidden representation to a 1-hot encoding of the category
\end{itemize}

The learnable components of the decoding mechanism are intentionally kept simple to encourage the model to capture complexity within the representations themselves, rather than in the decoding functions.

\paragraph{Pre-Training Procedure}
Our goal is to enable the foundation model to learn relationships between features. To this end, we adopt a self-supervised approach (requiring no annotations) based on masked token prediction, similar to the pretraining strategy used in language models such as BERT~\cite{devlin-etal-2019-bert}.
While training using \emph{random masking} is common in language models, communication data requires more care. At each position in the input sequence, multiple features are present, and some features are not predictable from others. Naively masking such features can introduce instability during training.

To address this, we identify a subset of five \emph{target} features from Table~\ref{table:feature-summary}—namely, the transmission rank, selected precoder, Doppler spectrum, center and length of the delay profile. These are features that can be estimated from the remaining ones, as discussed in Section \ref{sec:comms}. Pretraining is then performed by randomly masking a subset of the target features at each slot in the input sequence and training the model to predict them. An overview of the model and pre-training procedure is provided in Figure~\ref{fig:model}.

\section{Experiments}
The evaluation of our foundation model focuses on demonstrating its effectiveness in understanding communication systems data and its ability to perform estimation tasks. Specifically, we consider two scenarios:
\begin{itemize}
    \item[$1.$] \textbf{Forecasting.} The model is tasked with estimating the value of a given feature in the next slot.
    \item[$2.$] \textbf{Interpolation.} We randomly mask the values of a feature at certain slots in the input sequence, and the model is tasked with estimating the values of the masked feature. In more detail, we perform one evaluation run for each feature. During each run, for each batch, a slot of the input sequence is selected randomly (with uniform distribution), and the selected feature is masked for that slot.
\end{itemize}

\subsection{Experimental Setting}
\textbf{Data.} We generate 1 million datapoints using the simulator described in Section~\ref{sec:dataset}. The data is then randomly split into training, validation, and test sets, containing 80\%, 10\%, and 10\% of the data, respectively.

\textbf{Foundation Model Evaluation Procedure.} As described above, we evaluate the model by measuring its ability to estimate missing values for features of interest through the tasks of forecasting and interpolation. The features used in these tasks are the target features defined in Section~\ref{sec:method}: transmission rank, selected precoder, Doppler spread, center and length of the delay profile.

\textbf{Hyperparameters.} We use the validation set to tune the hyperparameters of our model. Specifically, we perform a grid search over learning rate, learning rate scheduling, token dimension, number of layers, and number of attention heads. 
We provide the values used for the hyperparameter tuning procedure in Table \ref{tab:hyp}.

\begin{table}[ht]
    \centering
    \caption{Values for hyperparameter tuning procedure.}
    \label{tab:hyp}
    \resizebox{\columnwidth}{!}{%
    \begin{tabular}{cc}
        \toprule
        \textbf{Feature} & \textbf{Values} \\
        \midrule
        Learning Rate & $10^{-4}, 10^{-5}, 10^{-6}$\\
        Learning Rate Schedule & None, Step, Cosine Annealing \\
        Token Dimension & 128, 256, 512\\
        Number of Layers & 8, 12, 16\\
        Number of Attention Heads & 4, 8, 16, 32\\
        \bottomrule
    \end{tabular}
    }
\end{table}

The final hyperparameter values used for our experiments are shown in Table \ref{tab:param_values}. During pre-training, we randomly (with uniform distribution) mask one target feature at each input slot. We use a batch size of 256, a learning rate of $10^{-6}$, and no learning rate schedule for all model and dataset sizes. 
\begin{table}[ht]
\caption{Parameters and dimensions for different model sizes used in our experiments.}
\label{tab:param_values}
\centering
\begin{tabular}{l|llll} 
\hline
Model & \#layers & \#head & \multicolumn{1}{c}{token dim} & hidden dim  \\ 
\hline
5M    & 6        & 6      & 387                           & 1548        \\
30M   & 32       & 18     & 387                           & 1548        \\
100M  & 75       & 24     & 429                           & 1716        \\
\hline
\end{tabular}
\end{table}

\textbf{Computational Resources.} For data generation we found that it takes 0.5 seconds to generate a single slot. We parallelize the data generation over multiple Intel Xeon Platinum processing nodes. For training our largest model, we use 2$\times$ NVIDIA A6000 GPUs.

\subsection{Results}
Table~\ref{tab:forecasting_results} presents the results for the \emph{forecasting} and \emph{interpolation} tasks. The model achieves consistently low mean squared error (MSE) across all five target features, demonstrating its ability to capture dependencies in the communication data. Since all features are normalized prior to training, the reported MSE values are computed on standardized scales and can be interpreted as percentage errors. As expected, interpolation performs slightly better than forecasting, as it estimates missing features using other available features within the same slot, while forecasting requires predicting feature values at future slots.


\begin{table}[ht]
\caption{Estimation Error (MSE) for the \textbf{forecasting} and \textbf{interpolation} tasks on five key features: center of the delay profile ($\hat{\mu}$), length of the delay profile ($\hat{\ell}$), Doppler spectrum width ($\hat{w}$), transmission rank ($\hat{R}$), and selected precoder ($\hat{\mathbf{W}}$).}
\vspace{2mm}
\centering
\resizebox{\columnwidth}{!}{%
\label{tab:forecasting_results}
\begin{tabular}{l|c:c:c:c:c}
~ & $\hat{\mu}$ & $\hat{\ell}$ & $\hat{w}$ & $\hat{R}$ & $\hat{\mathbf{W}}$ \\
\hline
\textbf{Forecasting}   & 0.019 & 0.021 & 0.077 & 0.129 & 0.101 \\
\textbf{Interpolation} & 0.019 & 0.020 & 0.054 & 0.128 & 0.100 \\

\end{tabular}
}
\end{table}

\begin{figure}[htbp]
\centerline{\includegraphics[width=0.8\columnwidth]{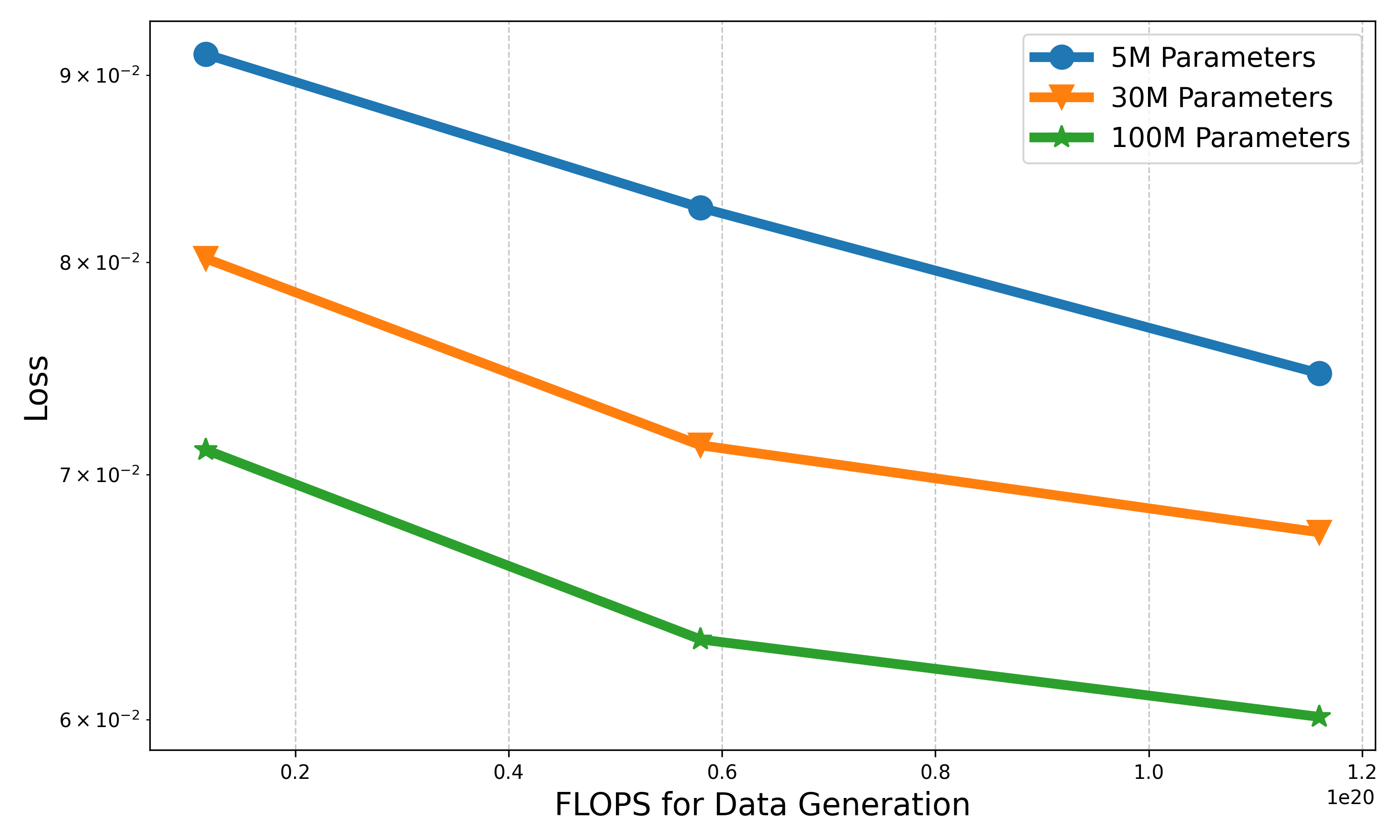}}
\vspace{-0.1cm}
\caption{Scaling behaviour vs. data generation compute. Each curve shows the test loss for a model of a given size (5M, 30M, or 100M parameters) as a function of compute used for data generation.}
\label{fig:scaling_data_gen}
\end{figure}

Among the features, Doppler spectrum width ($\hat{w}$) and delay profile parameters ($\hat{\mu}$, $\hat{\ell}$), which are all scalar-valued, are estimated with the highest accuracy. In contrast, transmission rank ($\hat{R}$) and especially the selected precoder ($\hat{\mathbf{W}}$)—a matrix-valued feature with variable dimensionality—exhibit higher errors. This reflects the increased difficulty of modeling high-dimensional, structured, and variable-size features. These observations are consistent with the challenges discussed in Section~IV and highlight the model’s ability to generalize across diverse modalities in communication systems data.

\subsection{Scaling Behaviour.}
We carried out experiments to study how model performance scales with compute, measured in floating point operations (FLOPs) allocated to data generation and training. Specifically, we trained foundation models of three sizes (5M, 30M, and 100M parameters) on datasets of varying sizes ($10^6$, $5 \cdot 10^6$, and $10^7$ training examples). These datasets are obtained by randomly subsampling the training set used for the estimation experiments. Each configuration was trained for up to 20 epochs.

\begin{figure}[htbp]
\centerline{\includegraphics[width=0.8\columnwidth]{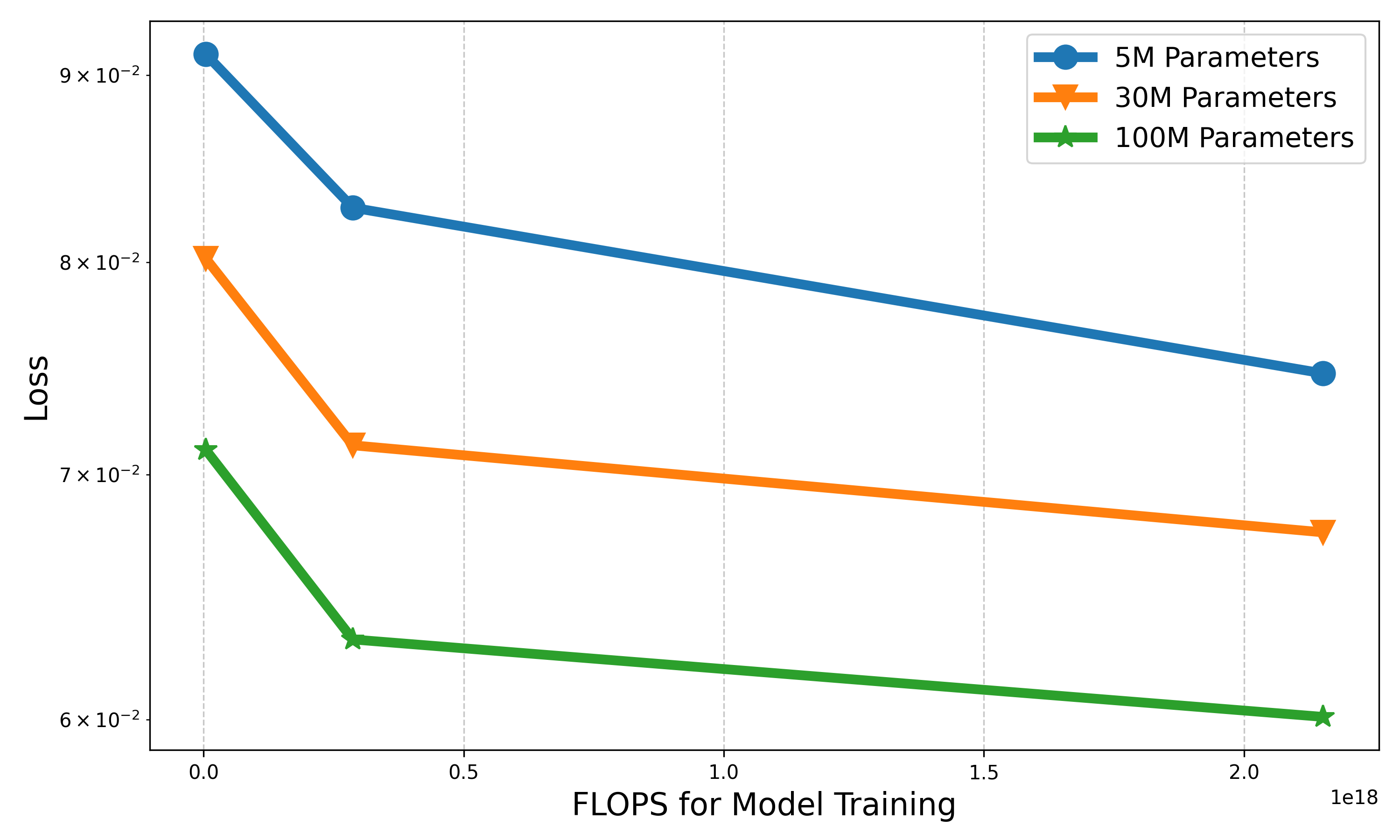}}
\vspace{-0.1cm}
\caption{Scaling behaviour vs. training compute. Each curve shows the test loss for models of a given size (5M, 30M, 100M) as a function of compute used for training.}
\label{fig:scaling_training}
\end{figure}

Figure~\ref{fig:scaling_data_gen} shows how the test loss (sum of the losses for the target features) varies with compute allocated to data generation, which determines dataset size. Figure~\ref{fig:scaling_training} shows test loss as a function of training compute, which controls the number of model updates. Performance improves with both model and dataset size, especially when scaled together. These trends are consistent with known scaling behaviour observed in other domains. Larger models benefit more from extended training. The results highlight the importance of co-scaling model size, data, and training budget to fully realize performance gains.

\section{Conclusion}
In this work, we laid the groundwork for a foundation model for communication systems. We discussed and addressed the unique challenges involved in designing such a model, developed a simulation system to generate training data, and demonstrated the model’s ability to perform forecasting and interpolation across several key features.

Future work will focus on scaling up both the dataset and model size, and incorporating additional features. We will release our data generation pipeline and model upon acceptance of this paper, with the hope that this work serves as a first step in this paradigm shift. We aim to support the community’s progress toward building datasets, models, and benchmarks that advance the development of effective foundation models for communication systems, with potential applications in 6G.




\section*{Impact Statement}
This paper presents work whose goal is to advance the field of 
Machine Learning. There are many potential societal consequences 
of our work, none which we feel must be specifically highlighted here.

\bibliography{example_paper}

\begin{thebibliography}{37}
\providecommand{\natexlab}[1]{#1}
\providecommand{\url}[1]{\texttt{#1}}
\expandafter\ifx\csname urlstyle\endcsname\relax
  \providecommand{\doi}[1]{doi: #1}\else
  \providecommand{\doi}{doi: \begingroup \urlstyle{rm}\Url}\fi

\bibitem[3GPP(2018)]{38211}
3GPP.
\newblock \emph{NR; Physical Channels and Modulation, 3GPP Standard TS 38.211},
  2018.

\bibitem[Abbaspourazad et~al.(2024)]{abbaspourazad2024largescale}
Abbaspourazad, S. et~al.
\newblock Large-scale training of foundation models for wearable biosignals.
\newblock In \emph{International Conference on Learning Representations}, 2024.
\newblock URL \url{https://openreview.net/forum?id=pC3WJHf51j}.

\bibitem[Alayrac et~al.(2022)]{alayrac2022flamingo}
Alayrac, J.-B. et~al.
\newblock Flamingo: a visual language model for few-shot learning.
\newblock In \emph{Advances in Neural Information Processing Systems}, 2022.
\newblock URL \url{https://openreview.net/forum?id=EbMuimAbPbs}.

\bibitem[Bodnar et~al.(2024)]{bodnar2024foundationmodelearth}
Bodnar, C. et~al.
\newblock A foundation model for the earth system.
\newblock \emph{arXiv}, 2024.
\newblock URL \url{https://arxiv.org/abs/2405.13063}.
\newblock \textit{arxiv:2405.13063}.

\bibitem[Bommasani et~al.(2021)]{Bommasani2021FoundationModels}
Bommasani, R. et~al.
\newblock On the opportunities and risks of foundation models.
\newblock \emph{arxiv}, 2021.
\newblock URL \url{https://arxiv.org/abs/2108.07258}.
\newblock \textit{arxiv:2108.07258}.

\bibitem[Bornea et~al.(2024)Bornea, Ayed, Domenico, Piovesan, and
  Maatouk]{bornea2024telcoragnavigatingchallengesretrievalaugmented}
Bornea, A.-L., Ayed, F., Domenico, A.~D., Piovesan, N., and Maatouk, A.
\newblock Telco-rag: Navigating the challenges of retrieval-augmented language
  models for telecommunications.
\newblock \emph{arxiv}, 2024.
\newblock URL \url{https://arxiv.org/abs/2404.15939}.
\newblock \textit{arxiv:2404.15939}.

\bibitem[Chen et~al.(2025)]{chen2025expandingperformanceboundariesopensource}
Chen, Z. et~al.
\newblock Expanding performance boundaries of open-source multimodal models
  with model, data, and test-time scaling.
\newblock \emph{arxiv}, 2025.
\newblock URL \url{https://arxiv.org/abs/2412.05271}.
\newblock \textit{arxiv:2412.05271}.

\bibitem[Dai et~al.(2024)]{dai2024nvlmopenfrontierclassmultimodal}
Dai, W. et~al.
\newblock Nvlm: Open frontier-class multimodal {LLMs}.
\newblock \emph{arxiv}, 2024.
\newblock URL \url{https://arxiv.org/abs/2409.11402}.
\newblock \textit{arxiv:2409.11402}.

\bibitem[Das et~al.(2024)Das, Kong, Sen, and
  Zhou]{das2024decoderonlyfoundationmodeltimeseries}
Das, A., Kong, W., Sen, R., and Zhou, Y.
\newblock A decoder-only foundation model for time-series forecasting.
\newblock \emph{arxiv}, 2024.
\newblock URL \url{https://arxiv.org/abs/2310.10688}.
\newblock \textit{arxiv:2310.10688}.

\bibitem[Devlin et~al.(2019)Devlin, Chang, Lee, and
  Toutanova]{devlin-etal-2019-bert}
Devlin, J., Chang, M.-W., Lee, K., and Toutanova, K.
\newblock {BERT}: Pre-training of deep bidirectional transformers for language
  understanding.
\newblock In \emph{Conference of the North {A}merican Chapter of the
  Association for Computational Linguistics: Human Language Technologies}, pp.\
   4171--4186. Association for Computational Linguistics, 2019.
\newblock \doi{10.18653/v1/N19-1423}.
\newblock URL \url{https://aclanthology.org/N19-1423/}.

\bibitem[Dosovitskiyet et~al.(2021)]{dosovitskiy2021an}
Dosovitskiyet, A. et~al.
\newblock An image is worth 16x16 words: Transformers for image recognition at
  scale.
\newblock In \emph{International Conference on Learning Representations}, 2021.
\newblock URL \url{https://openreview.net/forum?id=YicbFdNTTy}.

\bibitem[Ekambaram et~al.(2024)]{ekambaram2024tiny}
Ekambaram, V. et~al.
\newblock Tiny time mixers ({TTM}s): Fast pre-trained models for enhanced
  zero/few-shot forecasting of multivariate time series.
\newblock In \emph{Conference on Neural Information Processing Systems}, 2024.
\newblock URL \url{https://openreview.net/forum?id=3O5YCEWETq}.

\bibitem[Goswami et~al.(2024)Goswami, Szafer, Choudhry, Cai, Li, and
  Dubrawski]{goswami2024moment}
Goswami, M., Szafer, K., Choudhry, A., Cai, Y., Li, S., and Dubrawski, A.
\newblock Moment: A family of open time-series foundation models.
\newblock In \emph{International Conference on Machine Learning}, 2024.

\bibitem[Grattafiori et~al.(2024)]{grattafiori2024llama3herdmodels}
Grattafiori, A. et~al.
\newblock The {Llama} 3 herd of models.
\newblock \emph{arxiv}, 2024.
\newblock URL \url{https://arxiv.org/abs/2407.21783}.
\newblock \textit{arxiv:2407.21783}.

\bibitem[Hamdan et~al.(2023)]{s23218792}
Hamdan, M.~Q. et~al.
\newblock Recent advances in machine learning for network automation in the
  o-ran.
\newblock \emph{Sensors}, 23\penalty0 (21), 2023.
\newblock ISSN 1424-8220.
\newblock \doi{10.3390/s23218792}.
\newblock URL \url{https://www.mdpi.com/1424-8220/23/21/8792}.

\bibitem[Hoydis et~al.(2022)]{sionna}
Hoydis, J. et~al.
\newblock Sionna: An open-source library for next-generation physical layer
  research.
\newblock \emph{arxiv}, 2022.
\newblock URL \url{https://arxiv.org/abs/2203.11854}.
\newblock \textit{arxiv:2203.118548}.

\bibitem[Lewis et~al.(2020)]{10.5555/3495724.3496517}
Lewis, P. et~al.
\newblock Retrieval-augmented generation for knowledge-intensive nlp tasks.
\newblock In \emph{Proceedings of the 34th International Conference on Neural
  Information Processing Systems}, NIPS '20, Red Hook, NY, USA, 2020. Curran
  Associates Inc.
\newblock ISBN 9781713829546.

\bibitem[Li et~al.(1998)Li, Cimini, and Sollenberger]{Li_robustCE}
Li, Y., Cimini, L., and Sollenberger, N.
\newblock Robust channel estimation for ofdm systems with rapid dispersive
  fading channels.
\newblock \emph{IEEE Transactions on Communications}, 46\penalty0 (7):\penalty0
  902--915, 1998.
\newblock \doi{10.1109/26.701317}.

\bibitem[Liao et~al.(2020)Liao, Wei, and Zou]{Liao_2020}
Liao, F., Wei, S., and Zou, S.
\newblock Deep learning methods in communication systems: A review.
\newblock \emph{Journal of Physics: Conference Series}, 1617\penalty0
  (1):\penalty0 012024, aug 2020.
\newblock \doi{10.1088/1742-6596/1617/1/012024}.
\newblock URL \url{https://dx.doi.org/10.1088/1742-6596/1617/1/012024}.

\bibitem[Liu et~al.(2024)]{liu2024moiraimoeempoweringtimeseries}
Liu, X. et~al.
\newblock Moirai-moe: Empowering time series foundation models with sparse
  mixture of experts.
\newblock \emph{arxiv}, 2024.
\newblock URL \url{https://arxiv.org/abs/2410.10469}.
\newblock \textit{arxiv:2410.10469}.

\bibitem[Liu et~al.(2025)]{liu2025sundialfamilyhighlycapable}
Liu, Y. et~al.
\newblock Sundial: A family of highly capable time series foundation models.
\newblock \emph{arxiv}, 2025.
\newblock URL \url{https://arxiv.org/abs/2502.00816}.
\newblock \textit{arxiv:2502.00816}.

\bibitem[Narayanswamy et~al.(2025)]{narayanswamy2025scaling}
Narayanswamy, G. et~al.
\newblock Scaling wearable foundation models.
\newblock In \emph{International Conference on Learning Representations}, 2025.
\newblock URL \url{https://openreview.net/forum?id=yb4QE6b22f}.

\bibitem[Nikbakht et~al.(2024)Nikbakht, Benzaghta, and
  Geraci]{nikbakht2024tspecllmopensourcedatasetllm}
Nikbakht, R., Benzaghta, M., and Geraci, G.
\newblock Tspec-llm: An open-source dataset for llm understanding of 3gpp
  specifications.
\newblock \emph{arxiv}, 2024.
\newblock URL \url{https://arxiv.org/abs/2406.01768}.
\newblock \textit{arxiv:2406.1768}.

\bibitem[OpenAI(2024)]{openAIo1}
OpenAI.
\newblock Openai o1 system card, 2024.
\newblock \url{https://openai.com/index/openai-o1-system-card/}.

\bibitem[Piovesan et~al.(2024)Piovesan, Domenico, and
  Ayed]{piovesan2024telecomlanguagemodelslarge}
Piovesan, N., Domenico, A.~D., and Ayed, F.
\newblock Telecom language models: Must they be large?
\newblock \emph{arxiv}, 2024.
\newblock URL \url{https://arxiv.org/abs/2403.04666}.
\newblock \textit{arxiv:2403.04666}.

\bibitem[Rasul et~al.(2024)]{rasul2024lagllama}
Rasul, K. et~al.
\newblock {Lag-Llama}: Towards foundation models for probabilistic time series
  forecasting.
\newblock \emph{arxiv}, 2024.
\newblock \textit{arxiv:2310.08278}.

\bibitem[Saad et~al.(2024)]{saad2024artificialgeneralintelligenceaginative}
Saad, W. et~al.
\newblock Artificial general intelligence (agi)-native wireless systems: A
  journey beyond 6g.
\newblock \emph{arxiv}, 2024.
\newblock URL \url{https://arxiv.org/abs/2405.02336}.
\newblock \textit{arxiv:2405.02336}.

\bibitem[Shao et~al.(2024)]{10582827}
Shao, J. et~al.
\newblock Wirelessllm: Empowering large language models towards wireless
  intelligence.
\newblock \emph{Journal of Communications and Information Networks}, 9\penalty0
  (2):\penalty0 99--112, 2024.
\newblock \doi{10.23919/JCIN.2024.10582827}.

\bibitem[Shi et~al.(2025)Shi, Wang, Nie, Li, Ye, Wen, and Jin]{shi2025timemoe}
Shi, X., Wang, S., Nie, Y., Li, D., Ye, Z., Wen, Q., and Jin, M.
\newblock Time-moe: Billion-scale time series foundation models with mixture of
  experts.
\newblock In \emph{International Conference on Learning Representations}, 2025.
\newblock URL \url{https://openreview.net/forum?id=e1wDDFmlVu}.

\bibitem[Sun et~al.(2024)Sun, Lee, and Simpson]{s24061925}
Sun, Y., Lee, H., and Simpson, O.
\newblock Machine learning in communication systems and networks.
\newblock \emph{Sensors}, 24\penalty0 (6), 2024.
\newblock ISSN 1424-8220.
\newblock \doi{10.3390/s24061925}.
\newblock URL \url{https://www.mdpi.com/1424-8220/24/6/1925}.

\bibitem[Team et~al.(2024)]{geminiteam2024geminifamilyhighlycapable}
Team, G. et~al.
\newblock Gemini: A family of highly capable multimodal models.
\newblock \emph{arxiv}, 2024.
\newblock URL \url{https://arxiv.org/abs/2312.11805}.
\newblock \textit{arxiv:2312.11805}.

\bibitem[Thakkar et~al.(2020)Thakkar, Goyal, and Bhattacharyya]{THAKKAR2020119}
Thakkar, K., Goyal, A., and Bhattacharyya, B.
\newblock Emergence of deep learning as a potential solution for detection,
  recovery and de-noising of signals in communication systems.
\newblock \emph{International Journal of Intelligent Networks}, 1:\penalty0
  119--127, 2020.
\newblock ISSN 2666-6030.
\newblock \doi{https://doi.org/10.1016/j.ijin.2020.12.001}.
\newblock URL
  \url{https://www.sciencedirect.com/science/article/pii/S2666603020300270}.

\bibitem[Tse \& Viswanath(2005)Tse and Viswanath]{Tse_book}
Tse, D. and Viswanath, P.
\newblock \emph{Fundamentals of wireless communication}.
\newblock Cambridge University Press, USA, 2005.
\newblock ISBN 0521845270.

\bibitem[Wang \& Li(2024)Wang and
  Li]{wang2024machinelearningcommunicationsroad}
Wang, S. and Li, G.~Y.
\newblock Machine learning in communications: A road to intelligent
  transmission and processing.
\newblock \emph{arxiv}, 2024.
\newblock URL \url{https://arxiv.org/abs/2407.11595}.
\newblock \textit{arxiv:2407.11595}.

\bibitem[Wang et~al.(2023)]{10.1109/MWC.008.2200157}
Wang, Y. et~al.
\newblock Transformer-empowered 6g intelligent networks: From massive mimo
  processing to semantic communication.
\newblock \emph{Wireless Communications}, 30\penalty0 (6):\penalty0 127–135,
  December 2023.
\newblock ISSN 1536-1284.
\newblock \doi{10.1109/MWC.008.2200157}.
\newblock URL \url{https://doi.org/10.1109/MWC.008.2200157}.

\bibitem[Yu et~al.(2022)Yu, Sohrabi, and Jiang]{9914567}
Yu, W., Sohrabi, F., and Jiang, T.
\newblock Role of deep learning in wireless communications.
\newblock \emph{IEEE BITS the Information Theory Magazine}, 2\penalty0
  (2):\penalty0 56--72, 2022.
\newblock \doi{10.1109/MBITS.2022.3212978}.

\bibitem[Žeger \& Šišul(2021)Žeger and Šišul]{9550825}
Žeger, I. and Šišul, G.
\newblock Introduction to deep learning possibilities in communication systems.
\newblock In \emph{International Symposium ELMAR}, pp.\  21--24, 2021.
\newblock \doi{10.1109/ELMAR52657.2021.9550825}.

\end{thebibliography}
\bibliographystyle{icml2025}




\end{document}